\documentclass{ieeetj}
\usepackage[T1]{fontenc}
\usepackage{cite}
\usepackage{amsmath,amssymb,amsfonts}

\usepackage{graphicx,color}
\usepackage{textcomp}
\usepackage{xcolor}
\usepackage{booktabs}
\usepackage{array}
\newcolumntype{P}[1]{>{\raggedright\arraybackslash}p{#1}}
\usepackage{url}
\usepackage{tikz}
\usepackage{pgfplots}
\pgfplotsset{compat=1.18}
\usetikzlibrary{decorations.pathreplacing,arrows.meta,positioning}
\newcommand{\figsx}{\fontsize{6}{7}\selectfont}
\usepackage{needspace}
\usepackage{balance}
\usepackage{hyperref}
\hypersetup{hidelinks=true}
\newcommand{\Sstate}{\mathcal{S}_t}
\AtBeginDocument{\definecolor{tmlcncolor}{cmyk}{0.93,0.59,0.15,0.02}\definecolor{NavyBlue}{RGB}{0,86,125}}
\def\OJlogo{\vspace{-4pt}{\fontsize{7}{8}\selectfont Preprint, submitted to IEEE OJ-CS}}
\def\seclogo{\vspace{10pt}{\fontsize{7}{8}\selectfont Manuscript for review}}
\def\authorrefmark#1{\ensuremath{^{\textbf{#1}}}}

\usepackage{etoolbox}
\makeatletter
\patchcmd{\@maketitle}{\ifx\@revised\@empty}{\ifx\@reviseddate\@empty}{}{}
\patchcmd{\@maketitle}{\fi.\space}{\fi\space}{}{}
\def\ps@headings{%
  \def\@oddhead{\hbox to\textwidth{\scriptsize\sffamily Preprint, submitted to IEEE OJ-CS\hfil Xu et al.}}%
  \let\@evenhead\@oddhead
  \def\@oddfoot{\hbox to\textwidth{\hfil\scriptsize\thepage}}%
  \let\@evenfoot\@oddfoot}
\let\ps@plain\ps@headings
\makeatother

\begin{document}
\pagestyle{headings}
\receiveddate{}
\reviseddate{}
\accepteddate{}
\publisheddate{}
\currentdate{}
\doiinfo{}

\markboth{}{Xu \emph{et al.}}

\title{The State Is Not the Bottleneck: A Paged-State Gated-DeltaNet Tile on a Tri-Mode Multicast Mesh, Bit-Exact from RTL to Silicon}

\author{CONG XU\authorrefmark{1}, \IEEEmembership{Member, IEEE}, and RAVI SANKAR\authorrefmark{1}, \IEEEmembership{Life Senior Member, IEEE}}
\affil{iCONS Laboratory, Department of Electrical and Computer Engineering, University of South Florida, Tampa, FL 33620 USA}
\corresp{Corresponding author: Cong Xu (email: cong@usf.edu).}

\begin{abstract}
Linear-attention language models replace a growing key--value cache with a recurrent state of fixed size.
On an edge FPGA, this state can exceed the on-chip memory and must be paged to external memory.
This paper asks whether such paging dominates the cost of decoding (generating tokens).
We built a hardware tile that computes one complete layer of a custom 369-million-parameter hybrid Gated DeltaNet model on an Agilex 5 FPGA and pages its recurrent state.
The model state needs 6.2 MiB, whereas the device has 0.87 MiB of block memory.
Paging adds 26.0 MB of memory traffic per decoded token, at most 11.2 percent of the calculated total.
On the board at 75 MHz, with synthetic weights, one token takes 11.15 ms in this layer.
The matrix-vector products with the weights take about 62 percent of this time and the state phase, including the paging, 13 percent.
The tile is one node of a multicast mesh of tri-mode processing elements.
We measure the error of the fixed-point arithmetic against floating point and check that the hardware agrees bit for bit with a fixed-point reference.
In simulation, the state logic agrees with the reference on all 18 recurrent layers.
On the board, six builds pass all 30 runs of 64 tokens and two runs of 1,024 tokens.
For this model and memory interface, the delivery of the weights limits the decoding performance, and the recurrent state does not.
State paging is therefore practical for such a model on an edge FPGA.

\end{abstract}

\begin{IEEEkeywords}Linear attention, Gated DeltaNet, state paging, FPGA, network-on-chip, fixed-point verification.\end{IEEEkeywords}

\maketitle

\section{Introduction}
\IEEEPARstart{L}{inear} attention replaces the growing key--value (KV) cache of a transformer language model with a recurrent state of fixed size~\cite{katharopoulos2020linear,yang2025gdn}.
This state can still be larger than the on-chip memory of an edge FPGA.
Our custom model, a hybrid of the Gated DeltaNet type~\cite{yang2025gdn}, has 369 million parameters and 6.2~MiB of recurrent state, whereas our target device, an Agilex~5 FPGA, has 0.87~MiB of block memory.
The state must therefore reside in external memory, and the hardware must page it in and out for every token.
Our first research question is whether this paging dominates the cost of decoding, that is, of generating one token at a time.
We answer it in hardware, for one complete layer of the model.

\begin{figure}[t]
\centering
\begin{tikzpicture}[font=\figsx, node distance=1.2mm,
  b/.style={draw, rounded corners=1.2pt, align=center, inner sep=2pt, minimum height=6.5mm, text width=14.5mm},
  ar/.style={-{Latex[length=1.3mm]}}]
\node[b] (m) {56-mic array\\+ camera};
\node[b, right=of m] (f) {acoustic front-end\\fixed point, 7.02 ms/frame};
\node[b, right=of f, fill=black!6] (c) {scene cache\\$\Sstate$, $\eta{=}0.90$};
\node[b, right=2.4mm of c, fill=black!6] (d) {decode tile\\one model layer, LPDDR4};
\node[b, right=of d] (o) {advisory\\to planner};
\draw[ar] (m) -- (f); \draw[ar] (f) -- (c); \draw[ar, <->] (c) -- (d); \draw[ar] (d) -- (o);
\end{tikzpicture}\\[2pt]
\begin{minipage}[b]{0.52\columnwidth}\centering
\begin{tikzpicture}
\node[anchor=south west, inner sep=0] (ph) at (0,0) {\includegraphics[width=\linewidth]{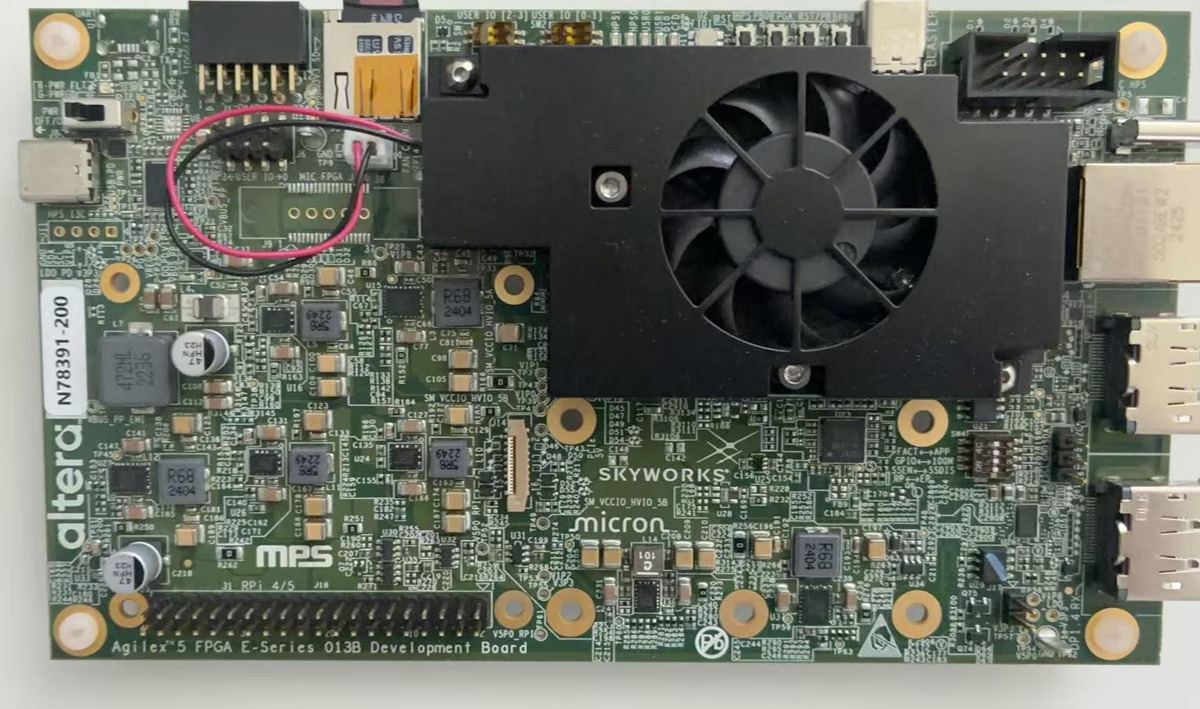}};
\begin{scope}[x={(ph.south east)}, y={(ph.north west)}]
\tikzset{mk/.style={circle, draw=white, fill=black!80, text=white, inner sep=0.2pt, minimum size=2.6mm, font=\figsx\bfseries}}
\node[mk] at (0.64,0.70) {1};
\node[mk] at (0.745,0.95) {2};
\node[mk] at (0.03,0.76) {3};
\end{scope}
\end{tikzpicture}\\[1pt]
{\figsx (a) The kit as used: (1) FPGA and the LPDDR4 memory of its programmable logic, under the heatsink, (2) on-board USB-Blaster~III, (3) USB-PD power.\par}
\end{minipage}\hfill
\begin{minipage}[b]{0.45\columnwidth}\centering
\begin{tikzpicture}[font=\figsx, node distance=4.8mm,
  bx/.style={draw, rounded corners=1pt, align=center, inner sep=1.6pt, text width=30mm},
  hd/.style={bx, fill=black!8},
  ar/.style={-{Latex[length=1.2mm]}}, lb/.style={font=\figsx, fill=white, inner sep=0.6pt}]
\node[bx] (host) {host PC: Python test scripts\\reference model, readback, checks};
\node[bx, below=of host] (fab) {Agilex 5 E-Series 013B, programmable logic:\\front-end and cache (separate build)\\decode tile, alone or on the $2\times3$ mesh\\hard processor system idle};
\node[hd, below=of fab] (emif) {LPDDR4 controller (hard IP)\\Avalon-MM $\to$ AXI4 bridges};
\node[hd, below=of emif] (dram) {LPDDR4 memory, 2~GB:\\INT4 weights and state pages};
\draw[ar, <->] (host) -- node[lb] {USB-Blaster~III, virtual JTAG} (fab);
\draw[ar, <->] (fab) -- node[lb] {512-bit ports} (emif);
\draw[ar, <->] (emif) -- node[lb] {x32, 2133~MT/s, 8.53~GB/s peak} (dram);
\end{tikzpicture}\\[1pt]
{\figsx (b) Where each part runs.\par}
\end{minipage}
\caption{The application and the platform. Top: the multi-rate architecture. The acoustic front-end detects a hidden vehicle, the cache answers queries about an unchanged scene with the hit rate $\eta$, and a cache miss invokes the language model, of which the decode tile implements one recurrent layer. (a) The DK-A5E013BM16AEA development kit, which we used for all measurements. (b) Where each part runs. The host reaches the programmable logic only through JTAG, and the tile reaches its weights and its state pages only through the LPDDR4 memory.}
\label{fig:app}
\end{figure}

\looseness=-1 The model is intended for the following application, which Fig.~\ref{fig:app} shows together with the hardware platform.
At a blind junction, a microphone array can detect an approaching vehicle before a camera sees it~\cite{schulz2021hearing}.
Our companion work, LVC-World, stores such detections in a structured description of the scene~\cite{xu2026lvcworld}.
A language-model stage, which we refer to as the \emph{semantic layer}, answers semantic queries about the scene, but it is much slower than the acoustic detection.
The system therefore stores the answers in a scene cache and invokes the semantic layer again only when the scene changes, which is a cache miss.
This paper presents hardware for the language model of the semantic layer.
This hardware, which we refer to as the \emph{decode tile}, computes one layer of the model for one token.
We evaluated the tile and the acoustic processing of the application (the front-end) on the same FPGA, in separate bitstreams (configurations of the FPGA).

\looseness=-1 The design point is one tile that computes the recurrent layers of the model one after another.
In such a system, the weights and the recurrent state of all layers reside in external memory, so the tile reads the weights of the current layer from there and pages its state in and out.
Our experiments exercise the tile with one layer.
The closest related accelerator, in contrast, keeps a recurrent state of 2~MB on chip~\cite{gupta2026persistent}.
Two calculated numbers indicate that the state is a small part of the work of our model.
First, for an input text (prompt) of 744 tokens, the update of the recurrent state is only 2.48\,\% of the multiply-accumulate operations (MACs) per decoded token.
Second, the paging moves 26.0~MB per decoded token, which is 7.7\,\% or 11.2\,\% of the calculated memory traffic for the two number formats that we consider (Section~\ref{sec:slowrate}-\ref{sec:traffic}).
For both formats, the weights are the largest part of the traffic.

\looseness=-1 Measurements on the board support this calculation.
We report the latency per \emph{layer-token}, which is one token passing through one layer.
At 75~MHz and with synthetic weights, the tile takes 11.15~ms per layer-token.
The matrix-vector products with the weights take about 62\,\% of the cycles, the other arithmetic of the layer takes 25\,\%, and the state phase, which includes the paging, takes 13\,\%.
For this model and this memory interface, the state is therefore not the bottleneck.

\looseness=-1 Our second research question is whether the hardware computes exactly what it should.
A recurrent state carries a numerical error from one token to all later tokens, so we require bit-exact agreement between the hardware and its reference.
In this paper, \emph{bit-exact} means that a hardware result equals the result of a fixed-point software reference in every bit.
We check this agreement in the simulation of the register-transfer-level (RTL) design and on the FPGA board, which the title refers to as silicon.
Section~\ref{sec:slowrate}-\ref{sec:verify} also evaluates the accuracy of the fixed-point reference against 32-bit floating point (fp32).

\subsection{Scope and contributions}

\looseness=-1 The decode tile implements one complete recurrent layer of the model, which consists of input and output projections, a causal convolution, a gated update of the recurrent state and a gated multilayer perceptron (MLP).
Of the 24 layers of the model, 18 are of this kind.
The other six layers are attention layers, and they are outside the scope of this paper, as is the complete inference loop of the model.
The tile is one node of a mesh network-on-chip.
The other five nodes are processing elements that switch among three operating modes (tri-mode), and one request can address several nodes (multicast).
On the board, the complete layer runs with synthetic weights, which have the memory layout of the real weights and for which the simulation provides the exact expected values, and a separate test runs the update of the state with real inputs.

The paper makes three contributions.
\begin{itemize}
\item It quantifies the cost of state paging for a complete layer, by calculation (Section~\ref{sec:slowrate}) and by measurement on an edge FPGA (Section~\ref{sec:board}).
\item It integrates the tile with a mesh of tri-mode processing elements and measures the cost of a mode switch in clock cycles (Sections~\ref{sec:substrate} and~\ref{sec:board}).
\item It presents a verification flow that measures the error of the fixed-point arithmetic against fp32 in software and checks the RTL and the board for bit-exact agreement with the fixed-point reference (Sections~\ref{sec:slowrate}-\ref{sec:verify} and~\ref{sec:board}).
This flow exposed an error in a numerical limit of the RTL and several implementation faults that the original simulation tests had not detected.
\end{itemize}

\section{Related Work}\label{sec:related}

\subsection{Linear attention and its accelerators}

\looseness=-1 Linear attention expresses decoding as a recurrence over a matrix-valued state~\cite{katharopoulos2020linear}.
The delta rule adds targeted updates to this recurrence~\cite{schlag2021fwp}, and data-dependent gates control how fast the state decays~\cite{yang2024gla}.
Structured state-space duality relates state-space models to attention variants~\cite{dao2024ssd}.
Gated DeltaNet combines gating with the delta rule~\cite{yang2025gdn}.
Its state has $d_k\times d_v$ entries for each attention head, where $d_k$ and $d_v$ are the dimensions of the key and of the value, and this size does not depend on the length of the sequence.

\looseness=-1 Prior FPGA work on transformer decoding includes multi-FPGA appliances~\cite{hong2022dfx}, spatial-dataflow analysis~\cite{chen2024spatial}, bandwidth-aware mapping~\cite{zeng2024flightllm} and the edge accelerators TeLLMe and EdgeLLM~\cite{tellme2025,edgellm2024}.
All of these designs manage a KV cache, not a recurrent matrix state.
Several accelerators target Mamba, which is another recurrent architecture~\cite{emamba2025,mambax2025,wei2025lightmamba,marca2024,zhong2025specmamba,xu2026lowrankssm}.
None of them evaluates the paging of a Gated DeltaNet state, which we study here.
For its number formats, our tile uses two known techniques.
Weight-only quantization reduces the storage of the model~\cite{frantar2023gptq,lin2024awq}, and shared-exponent formats allow arithmetic with narrow mantissas~\cite{rouhani2020msfp}.

\subsection{Comparison with the closest prior work}

\looseness=-1 Gupta \emph{et al.} evaluate the recurrent update of Gated DeltaNet on an Alveo U55C FPGA card, with a state that stays on chip~\cite{gupta2026persistent}.
We refer to this update as the \emph{recurrent kernel}, and we adopt their fused formulation of it.
Our work differs in two respects.
First, their 2~MB state stays in block memory, whereas the recurrent state of our model exceeds the block memory of our device.
Second, they evaluate the kernel alone, whereas our tile implements the complete layer with its projections, its convolution and its gated MLP.
The latency of their kernel and the latency of our layer are therefore not directly comparable.
A study of a 1.2B-parameter gated state-space model also finds that weight reads dominate the memory traffic of decoding~\cite{madangopal2026precog}, and DeltaLog reduces the write traffic of the recurrent state for cases in which maintaining the state is costly~\cite{lin2026deltalog}.

\looseness=-1 Other designs change the function of the hardware at run time.
PD-Swap uses dynamic partial reconfiguration to change between the processing of the prompt (prefill) and decoding, and it reports about 45~ms per reconfiguration~\cite{zhang2025pdswap}.
TRINE switches dataflows on a shared processing array after it drains the pipeline~\cite{oh2026trine}.
Our processing element has three fixed operating modes and changes its mode without loading a bitstream.
A mode switch takes three cycles when the element is idle (simulation) and at most 34 cycles when it must first finish its work (board test).
The three mechanisms differ in kind, so these numbers are given for orientation only.

\looseness=-1 Multicast by destination mask is known from accelerator interconnects~\cite{colagrande2025mcast}.
Our design adopts it and converts each multicast request into one unicast message per destination at the point of injection.
Our contribution here is the integration of multicast with the tile and the processing elements.

\looseness=-1 Exact comparison with a software reference is an established practice for arithmetic units~\cite{ieee754,hauser_softfloat}.
BitLogic also checks neural-network RTL for FPGAs against a Python reference in simulation~\cite{buhrer2026bitlogic}.
Our evaluation extends such comparisons from simulation to execution on the board.
Table~\ref{tab:prior} compares the closest prior work with this work on four criteria, namely what is implemented, where the recurrent state is stored, the device and the type of evidence, and the check of exactness.
The last row of the table summarizes this work, which implements a complete layer, pages the state and reports exact comparisons in simulation and on the board.

\begin{table*}[t]
\caption{Comparison with the closest prior work on the four criteria of this paper.
``Resident'' means that the recurrent state stays on chip. ``Paged'' means that the state is
read from and written to external memory in pages. Cells report what the cited works
state. A dash means that the work does not address that criterion.}
\label{tab:prior}
\centering
\footnotesize
\setlength{\tabcolsep}{5pt}
\begin{tabular}{@{}P{2.3cm}P{3.7cm}P{2.6cm}P{4.4cm}P{3.2cm}@{}}
\toprule
Work & What is implemented & Recurrent state & Device and evidence & Exactness \\
\midrule
Gupta \emph{et al.}~\cite{gupta2026persistent} & Gated DeltaNet decode kernel, five fused phases & 2\,MB, resident in block memory & Alveo U55C (datacenter). Cycle counts from high-level synthesis and power after implementation. No board run reported & --- \\
DUET~\cite{kanani2026duet} & Mamba--Transformer hybrid, prefill and decode packages & kept in package-local buffers, residency not stated & 28\,nm ASIC synthesis and a cycle-accurate simulator. No prototype & --- \\
Mamba accelerators~\cite{emamba2025,mambax2025,wei2025lightmamba,marca2024,zhong2025specmamba,xu2026lowrankssm} & Mamba layers & no comparable per-head paging evaluation & FPGA (LightMamba, SpecMamba, LowRank-SSM, eMamba) or ASIC synthesis (MARCA, Mamba-X) & --- \\
TRINE~\cite{oh2026trine} & shared processing array, switched among systolic, SIMD and adder-tree dataflows & --- & Alveo U50 and ZCU104 named. Switch cost reported as negligible in wall time, not in cycles & --- \\
BitLogic~\cite{buhrer2026bitlogic} & logic-gate networks & --- & FPGA, post-route figures. Verilator & bit-exact to its Python reference, in simulation \\
\midrule
This work & complete Gated DeltaNet layer of a 369M hybrid model & paged, 26.0\,MB per decoded token & Agilex~5 board. Six final builds pass 30 runs of 64 layer-tokens & state engine exact in 72 simulation runs. Kernel and complete layer checked on the board \\
\bottomrule
\end{tabular}
\end{table*}

\subsection{Embodied perception and acoustic front-ends}

\looseness=-1 Language models for embodied systems such as vehicles and robots~\cite{hwang2024emma,tian2024drivevlm,kim2024openvla} take a long time per step on edge GPUs compared with the period of a control loop~\cite{taherin2025}.
Prior work addresses this mismatch by compression~\cite{wen2024tinyvla}, by reuse~\cite{xu2025vlacache} and by the separation of fast and slow paths~\cite{sun2026unifs,williams2026litevla}.
Our scene cache takes a different route because it stores a description of the scene that does not depend on the model.
The acoustic processing that fills the cache early uses the generalized cross-correlation (GCC) and the steered response power with phase transform (SRP-PHAT)~\cite{knapp1976gcc,dibiase2001srp,grondin2019odas}.
It runs on the occluded-vehicle recordings of Schulz \emph{et al.}~\cite{schulz2021hearing}, which we refer to as OVAD.
Earlier FPGA systems for microphone arrays~\cite{dasilva2017fpga} and SRP-PHAT front-ends for the extreme edge~\cite{yin2023srpedge} established this kind of processing.
Our contribution is to use the acoustic detection to fill the scene cache before the vehicle becomes visible.

\section{System Context: The Multi-Rate Perception Architecture}\label{sec:invariants}

\looseness=-1 This section describes the perception system that the decode tile serves, and it derives how often the semantic layer must run.
The top row of Fig.~\ref{fig:app} shows the architecture, which has two parts that run at different rates.
The front-end processes every sensor frame at the sensor rate of 10--100~Hz, and the semantic layer runs only when the scene changes.
The front-end maintains a \emph{scene description} $\Sstate=\{\mathcal{E}_t,\mathcal{R}_t,\mathcal{C}_t,\mathcal{P}_t\}$, which stores entities, relations, context and predictive cues~\cite{xu2026lvcworld}.
The scene description supplies the planner and the controller, and it also serves as the scene cache for semantic queries.
When only positions change, the system reuses the cached context and cues, which is a cache hit.
A cache miss occurs on a cold start or when a change of the scene is confirmed, for example a new object class or an acoustic alarm that is raised or cleared, and each miss invokes the semantic layer.

\emph{Effective query rate.} Let $\eta$ be the hit rate of the cache, that is, the fraction of sensor frames that are served from the cached context $\mathcal{C}_t$ and cues $\mathcal{P}_t$.
The semantic layer then receives queries at the rate
\begin{equation}
f_{\mathrm{sem,eff}} = (1-\eta)\, f_\mathrm{sensor},
\label{eq:effrate}
\end{equation}
where $f_\mathrm{sensor}$ is the sensor rate.
A single serial server avoids a growing queue only if its mean service time, that is, the latency of one query in the semantic layer, is below the average interval between queries, $T_\mathrm{avg}=1/f_{\mathrm{sem,eff}}$.
At a sensor rate of 30~Hz, for example, $\eta=0.80$ gives $T_\mathrm{avg}=167$~ms, which is enough for a service time of 150~ms, and $\eta=0.90$ gives 333~ms.
We therefore use $\eta\approx0.80$ as the feasibility threshold in this example.
To measure $\eta$, we replayed the 83 recordings (\emph{clips}) of the OVAD static test set through the cache policy in software.
Over these clips, the mean $\eta$ is 0.929 and the minimum is 0.852, and the minimum corresponds to $T_\mathrm{avg}=225$~ms in the 30~Hz example.

\looseness=-1 The mean does not show whether the misses arrive in bursts.
We therefore passed the misses of this replay through a queue model of the semantic layer, which is a first-in-first-out server with a fixed service time.
The replay advances in steps of 100~ms (\emph{hops}), and it contains 510 misses in 7{,}333 hops.
The longest burst of misses is 2, and the queue never holds more than 3 queries.
After a miss, the cached result is stale until the query is served, and we refer to this interval as the \emph{blind window}.
With a service time of 333~ms, 84.9\,\% of the queries finish within 333~ms, and the longest blind window is 832~ms.
With a service time of 150~ms, every query finishes within 200~ms.
Both service times are chosen for illustration.

\looseness=-1 In summary, the semantic layer receives about one query for every 14 sensor frames at the mean hit rate of 0.929.

\looseness=-1 \emph{Front-end.} The front-end processes the 56 microphone channels in three stages, which compute 512-point FFTs, PHAT whitening~\cite{knapp1976gcc} and SRP beamforming over 30 azimuths~\cite{dibiase2001srp}.
The results of the front-end match, bit for bit, a software model of the same integer arithmetic.
A 16-bit input gives a quantization SNR of about 98~dB, whereas the acoustic SNR of the workload is 10--30~dB, so the quantization noise is at least 68~dB below the acoustic noise.
Over 1{,}064 one-second segments, the correlation between the fixed-point and the 64-bit floating-point SRP spectra has a median of 1.0000 and a minimum of 0.907.

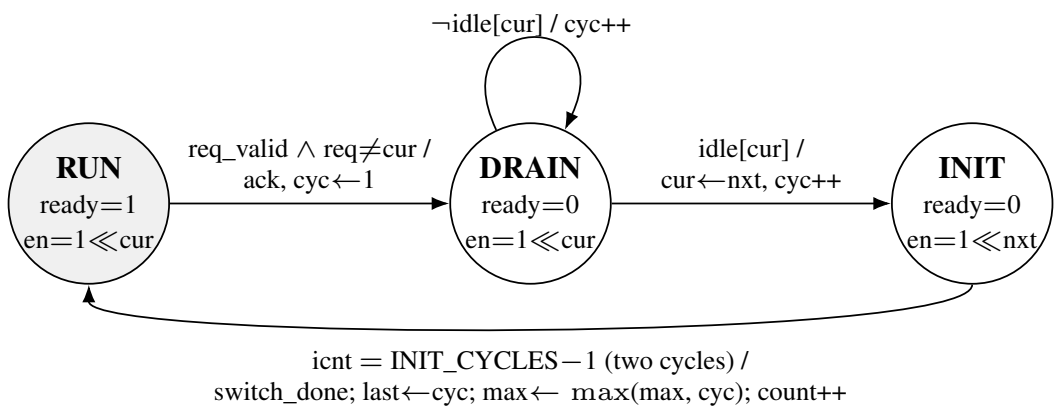
\begin{figure}[t]
\centering
\begin{minipage}[b]{\columnwidth}\centering
\resizebox{0.80\textwidth}{!}{%
\begin{tikzpicture}[font=\scriptsize,
  st/.style={draw, circle, minimum size=12.5mm, inner sep=0pt, align=center},
  ar/.style={-{Latex[length=1.4mm]}},
  cond/.style={font=\figsx, align=center, inner sep=1pt},
  sig/.style={font=\figsx, anchor=east, inner sep=1pt},
  wv/.style={line width=0.6pt},
  clk/.style={gray!45, line width=0.3pt, dotted}]

\node[st, fill=black!6] (run) at (1.0,0)
  {\textbf{RUN}\\{\figsx ready$=$1}\\{\figsx en$=$1$\ll$cur}};
\node[st] (drn) at (4.5,0)
  {\textbf{DRAIN}\\{\figsx ready$=$0}\\{\figsx en$=$1$\ll$cur}};
\node[st] (ini) at (8.0,0)
  {\textbf{INIT}\\{\figsx ready$=$0}\\{\figsx en$=$1$\ll$nxt}};
\draw[ar] (run) -- (drn) node[midway, above, cond, yshift=1pt]
  {req\_valid $\wedge$ req$\neq$cur /\\ack, cyc$\leftarrow$1};
\draw[ar] (drn) -- (ini) node[midway, above, cond, yshift=1pt]
  {idle[cur] /\\cur$\leftarrow$nxt, cyc{+}{+}};
\draw[ar] (drn) to[out=115, in=65, looseness=5]
  node[above, cond, yshift=-1pt] {$\neg$idle[cur] / cyc{+}{+}} (drn);
\draw[ar] (ini.south) .. controls (8.0,-1.1) and (1.0,-1.1) .. (run.south)
  node[pos=0.5, below, cond, yshift=-4pt]
  {icnt $=$ INIT\_CYCLES$-$1 (two cycles) /\\switch\_done; last$\leftarrow$cyc; max$\leftarrow\max$(max, cyc); count{+}{+}};

\end{tikzpicture}}
\end{minipage}
\caption{The mode controller of the processing element (Section~\ref{sec:substrate}). A mode switch costs $D+3$ cycles, where $D$ is the number of cycles that the current engine needs to become idle (drain). Arrow labels read ``condition / action''.}
\label{fig:pe}
\end{figure}

\section{Reconfigurable Tri-Mode Processing Element}\label{sec:substrate}

\looseness=-1 This section describes the tri-mode processing element, five of which surround the decode tile on the mesh.
An element has three engines and enables one of them at a time, which gives it three operating modes.
The parallel mode computes convolutions and dense inner products on 8-bit integers (INT8).
The sequential mode runs control tasks.
The memory-streaming mode buffers and moves data.
The engine of the parallel mode has 16 parallel arithmetic paths (\emph{lanes}).
Each lane computes a six-element INT8 dot product, so the engine performs 96 MACs per cycle with 18 DSP blocks, which are the hard multiplier blocks of the FPGA.

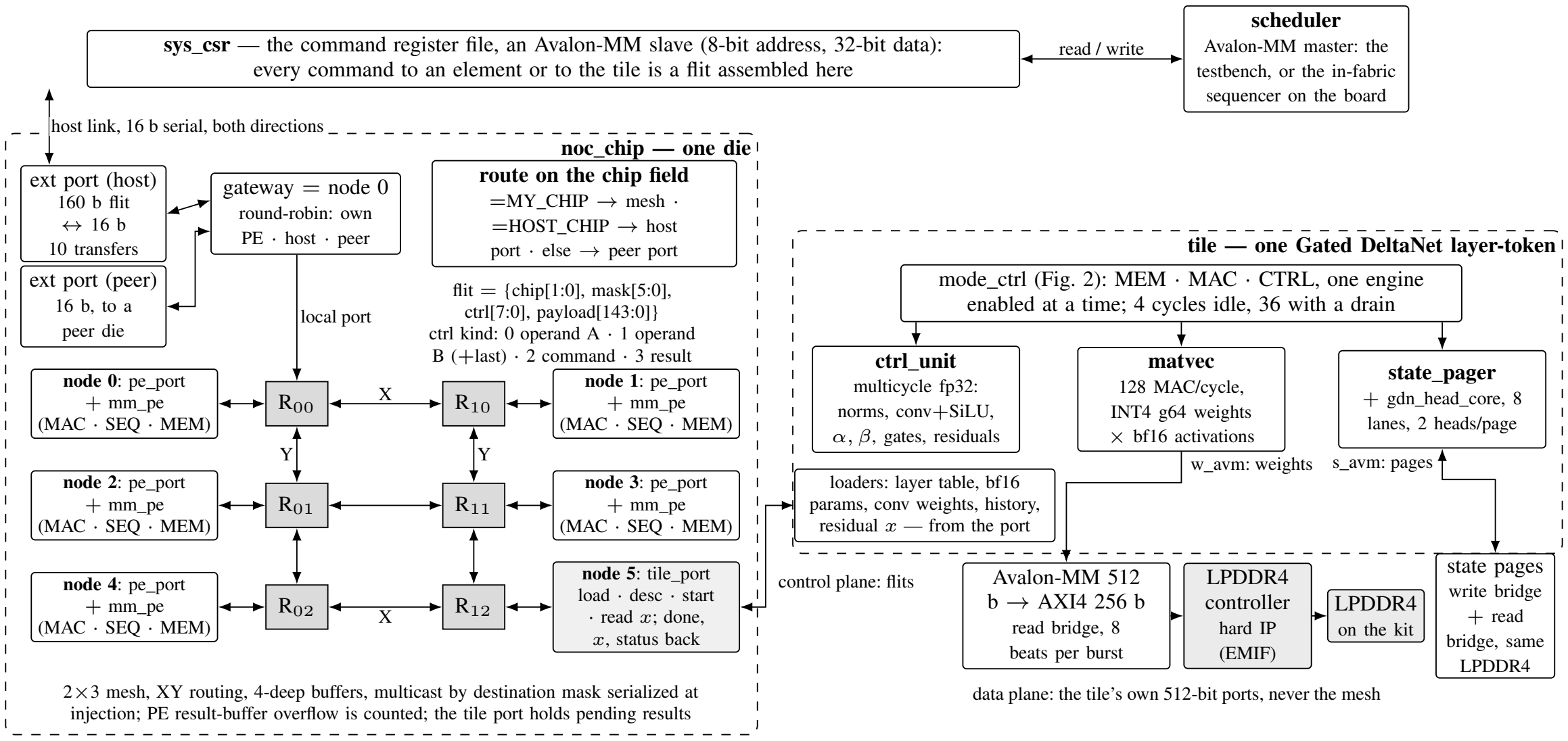
\begin{figure*}[t]
\centering
\resizebox{0.86\textwidth}{!}{%
\begin{tikzpicture}[font=\scriptsize,
  blk/.style={draw, rounded corners=1.2pt, align=center, inner sep=2pt},
  soft/.style={blk, fill=white},
  hard/.style={blk, fill=black!8},
  grp/.style={draw, dashed, rounded corners=3pt, inner sep=0pt},
  ttl/.style={font=\scriptsize\bfseries, inner sep=2pt},
  tn/.style={font=\figsx, align=center, inner sep=1pt},
  rt/.style={draw, fill=black!14, align=center, minimum width=7mm, minimum height=5mm, inner sep=1pt},
  pe/.style={blk, text width=20mm, inner sep=1.5pt, font=\figsx},
  ar/.style={-{Latex[length=1.4mm]}},
  bi/.style={{Latex[length=1.4mm]}-{Latex[length=1.4mm]}}]

\node[soft, text width=104mm] (csr) at (6.2,7.05)
  {\textbf{sys\_csr} --- the command register file, an Avalon-MM slave (8-bit address, 32-bit data):\\every command to an element or to the tile is a flit assembled here};
\node[soft, text width=24mm] (host) at (14.6,7.05)
  {\textbf{scheduler}\\{\figsx Avalon-MM master: the testbench, or the in-fabric sequencer on the board}};
\draw[bi] (csr.east) -- (host.west) node[midway, above, tn] {read / write};

\node[grp, minimum width=85mm, minimum height=68mm, anchor=south west] (noc) at (0,-0.6) {};
\node[ttl, anchor=north east] at (noc.north east) {noc\_chip --- one die};
\node[soft, text width=15mm] (hp) at (1.0,5.3)
  {ext port (host)\\{\figsx 160 b flit $\leftrightarrow$ 16 b\\10 transfers}};
\node[soft, text width=15mm] (pp) at (1.0,4.25)
  {ext port (peer)\\{\figsx 16 b, to a peer die}};
\node[soft, text width=20mm] (gw) at (3.4,5.3)
  {gateway $=$ node 0\\{\figsx round-robin: own PE $\cdot$ host $\cdot$ peer}};
\node[soft, text width=33mm] (dec) at (6.55,5.3)
  {\textbf{route on the chip field}\\{\figsx $=$MY\_CHIP $\to$ mesh $\cdot$ $=$HOST\_CHIP $\to$ host port $\cdot$ else $\to$ peer port}};
\node[tn, text width=42mm] at (6.3,4.05)
  {flit $=$ \{chip[1:0], mask[5:0], ctrl[7:0], payload[143:0]\}\\
   ctrl kind: 0 operand A $\cdot$ 1 operand B ($+$last) $\cdot$ 2 command $\cdot$ 3 result};
\draw[bi] (hp.east) -- (gw.west |- 0,5.45);
\draw[bi] (pp.east) -- (2.15,4.25) -- (2.15,5.1) -- (gw.west |- 0,5.1);
\node[rt] (r00) at (3.3,3.15) {R$_{00}$};
\node[rt] (r10) at (5.3,3.15) {R$_{10}$};
\node[rt] (r01) at (3.3,2.0)  {R$_{01}$};
\node[rt] (r11) at (5.3,2.0)  {R$_{11}$};
\node[rt] (r02) at (3.3,0.85) {R$_{02}$};
\node[rt] (r12) at (5.3,0.85) {R$_{12}$};
\draw[bi] (r00) -- (r10) node[midway, above, tn] {X};
\draw[bi] (r01) -- (r11); \draw[bi] (r02) -- (r12) node[midway, below, tn] {X};
\draw[bi] (r00) -- (r01) node[midway, left, tn] {Y}; \draw[bi] (r01) -- (r02);
\draw[bi] (r10) -- (r11) node[midway, right, tn] {Y}; \draw[bi] (r11) -- (r12);
\draw[ar] (gw.south -| r00) -- (r00.north) node[midway, right, tn] {local port};
\node[pe] (p0) at (1.35,3.15) {\textbf{node 0}: pe\_port $+$ mm\_pe\\(MAC $\cdot$ SEQ $\cdot$ MEM)};
\node[pe] (p2) at (1.35,2.0)  {\textbf{node 2}: pe\_port $+$ mm\_pe\\(MAC $\cdot$ SEQ $\cdot$ MEM)};
\node[pe] (p4) at (1.35,0.85) {\textbf{node 4}: pe\_port $+$ mm\_pe\\(MAC $\cdot$ SEQ $\cdot$ MEM)};
\node[pe] (p1) at (7.25,3.15) {\textbf{node 1}: pe\_port $+$ mm\_pe\\(MAC $\cdot$ SEQ $\cdot$ MEM)};
\node[pe] (p3) at (7.25,2.0)  {\textbf{node 3}: pe\_port $+$ mm\_pe\\(MAC $\cdot$ SEQ $\cdot$ MEM)};
\node[pe, fill=black!6] (tp) at (7.25,0.85) {\textbf{node 5}: tile\_port\\load $\cdot$ desc $\cdot$ start $\cdot$ read $x$; done, $x$, status back};
\draw[bi] (p0) -- (r00); \draw[bi] (p2) -- (r01); \draw[bi] (p4) -- (r02);
\draw[bi] (r10) -- (p1); \draw[bi] (r11) -- (p3); \draw[bi] (r12) -- (tp);
\node[tn, text width=80mm] at (4.25,-0.25)
  {2$\times$3 mesh, XY routing, 4-deep buffers, multicast by destination mask serialized at injection; PE result-buffer overflow is counted; the tile port holds pending results};
\draw[bi] (csr.south -| 0.5,0) -- (hp.north -| 0.5,0)
  node[pos=0.5, right, tn, fill=white, inner sep=0.5pt] {host link, 16 b serial, both directions};

\node[grp, minimum width=87mm, minimum height=36.5mm, anchor=south west] (tile) at (8.9,1.45) {};
\node[ttl, anchor=north east] at (tile.north east) {tile --- one Gated DeltaNet layer-token};
\node[soft, text width=62mm] (mc) at (13.3,4.4)
  {mode\_ctrl (Fig.~\ref{fig:pe}): MEM $\cdot$ MAC $\cdot$ CTRL, one engine enabled at a time; 4 cycles idle, 36 with a drain};
\node[soft, text width=22mm, minimum height=11mm] (cu) at (10.3,3.2)
  {\textbf{ctrl\_unit}\\{\figsx multicycle fp32: norms, conv$+$SiLU, $\alpha,\beta$, gates, residuals}};
\node[soft, text width=22mm, minimum height=11mm] (mv) at (13.3,3.2)
  {\textbf{matvec}\\{\figsx 128 MAC/cycle, INT4 g64 weights $\times$ bf16 activations}};
\node[soft, text width=22mm, minimum height=11mm] (pg) at (16.25,3.2)
  {\textbf{state\_pager}\\{\figsx $+$ gdn\_head\_core, 8 lanes, 2 heads/page}};
\draw[ar] (mc.south -| cu) -- (cu.north);
\draw[ar] (mc.south -| mv) -- (mv.north);
\draw[ar] (mc.south -| pg) -- (pg.north);
\node[soft, text width=28mm, font=\figsx] (ld) at (10.4,2.0)
  {loaders: layer table, bf16 params, conv weights, history, residual $x$ --- from the port};
\draw[bi] (tp.east) -- (8.6,0.85) |- (ld.west);
\node[tn, anchor=west, inner sep=0.5pt] at (8.72,1.12) {control plane: flits};

\node[soft, text width=22mm] (br) at (12.0,0.75)
  {Avalon-MM 512 b $\to$ AXI4 256 b\\{\figsx read bridge, 8 beats per burst}};
\node[hard, text width=13mm] (em) at (14.05,0.75) {LPDDR4 controller\\{\figsx hard IP (EMIF)}};
\node[hard, text width=9.5mm] (dr) at (15.5,0.75) {LPDDR4\\{\figsx on the kit}};
\node[soft, text width=12mm] (sr) at (16.85,0.75) {state pages\\{\figsx write bridge $+$ read bridge, same LPDDR4}};
\draw[ar] (br) -- (em); \draw[ar] (em) -- (dr);
\draw[ar] (mv.south) -- ++(0,-0.35) -| (br.north);
\node[tn, anchor=west, inner sep=0.5pt] at (13.38,2.45) {w\_avm: weights};
\draw[bi] (pg.south) -- ++(0,-0.35) -| (sr.north);
\node[tn, anchor=east, inner sep=0.5pt] at (16.17,2.45) {s\_avm: pages};
\node[tn, anchor=west] at (10.9,-0.15) {data plane: the tile's own 512-bit ports, never the mesh};
\end{tikzpicture}}
\caption{The implemented system. Left: the mesh, in which five nodes are tri-mode processing elements and node~5 is the decode tile. Right: the decode tile. The block names are those of the RTL: sys\_csr is the command register file, pe\_port with mm\_pe is a processing element, ctrl\_unit is the control datapath, matvec is the matrix-vector engine, and state\_pager with gdn\_head\_core is the state pager with the state engine. The mesh carries commands and results only, and the weights and the state pages use separate 512-bit ports to the LPDDR4 memory. The peer port can forward flits to a second system.}
\label{fig:sys}
\end{figure*}

\looseness=-1 Fig.~\ref{fig:pe} shows the mode controller, which has the states RUN, DRAIN and INIT.
A request to change the mode moves the controller from RUN to DRAIN.
There it lowers its ready signal and waits until the current engine is idle.
In INIT, it enables the target engine, waits two settling cycles and returns to RUN.
A mode switch therefore costs $D+3$ cycles, where $D$ is the number of cycles that the current engine needs to become idle.
In simulation, a switch from an idle element costs 3 cycles, and on the device a counter recorded a worst case of 34 cycles, or 0.34~$\mu$s at 100~MHz.
The decode tile has its own engines (Section~\ref{sec:slowrate}), and the same kind of controller selects among them.
The port of the tile adds a handshake state, so that a switch costs 4 cycles from idle and 36 cycles when an engine must first drain, both in simulation.

\looseness=-1 The elements and the tile communicate over a mesh network-on-chip, which carries messages of fixed size called \emph{flits}.
Each flit has a destination mask with one bit per node, so one request can address several nodes at once (multicast).
Fig.~\ref{fig:sys} shows the complete system, with the mesh of six nodes on the left, of which five are processing elements and node~5 is the decode tile, and the engines of the tile on the right.
The figure labels the parallel, sequential and memory-streaming modes of an element MAC, SEQ and MEM, respectively.
A scheduler sends commands to the nodes as flits and receives their results in the same way.
The mesh carries only these commands and results, whereas the weights and the state pages of the tile use two separate 512-bit ports to the memory.
An early design replicated multicast packets in the routers, and it deadlocked after 39 packets in simulation.
The final design sends the copies of a request one after another at the point of injection, which removes the dependency that we observed.
We have not proved that the mesh is free of deadlock in general.

\section{The Paged-State Decode Tile}
\label{sec:slowrate}

\looseness=-1 The decode tile is a block of hardware that computes one recurrent layer of the model for one token at a time, that is, with a batch size of one.
It consists of a matrix-vector engine, a state engine, a state pager and a control datapath (Fig.~\ref{fig:sys}, right).
A layer-token alternates among three kinds of phases.
In the matrix-vector phases, the tile reads weights from external memory and multiplies them by a vector.
In the state phase, it pages the recurrent state of the layer in, updates it and pages it out.
In the control phases, it computes the normalizations, the convolution and the gates.
The design point is one tile that serves all recurrent layers of the model in turn, so the tile keeps the weights and the state in external memory and not on chip.

\subsection{Computation of one layer}

A layer-token starts with a root-mean-square (RMS) normalization of the input vector.
Six input projections then produce the query $q$, the key $k$, the value $v$, the gate $z$ of the output normalization, and the inputs $a$ and $b$ of the decay and of the write strength, respectively.
The first three share one concatenated matrix, so four matrix products produce the six vectors.
A short causal convolution and a sigmoid linear unit (SiLU) activation process $q$, $k$ and $v$, and the query and the key are then normalized with the $\ell_2$ norm in each attention head.
After the update of the recurrent state, the layer applies a gated output normalization and an output projection.
A residual addition, a second normalization, a gated MLP and a second residual addition complete the layer.

\Needspace{5\baselineskip}
\looseness=-1 The update of the recurrent state, which is the recurrent kernel, follows the fused formulation of~\cite{gupta2026persistent}.
For one head with the state $S$, the decay $\alpha$, the write strength $\beta$ and the vectors $q$, $k$ and $v$ of the current token,
\begin{align}
s &= q\cdot k, &
r &= \alpha S^{\!\top} k, &
\hat o &= \alpha S^{\!\top} q,\\
\Delta v &= \beta\,(v - r), &
o &= \hat o + s\,\Delta v, &
S &\leftarrow \alpha S + k\,\Delta v^{\!\top}.
\end{align}
Here $S\in\mathbb{R}^{d_k\times d_v}$, $q,k\in\mathbb{R}^{d_k}$ and $v\in\mathbb{R}^{d_v}$, and $o$ is the output of the head.
The read pass computes the stored value $r$ and the decayed output $\hat o$ from the state, and the write pass applies the rank-one update of $S$.
These two passes dominate the cost of the kernel, but not the cost of the complete layer.

\looseness=-1 \emph{Model.} The model is a custom pruned and distilled derivative of Qwen3.5-0.8B-Base~\cite{qwen2026base}.
The hidden dimension remains 1{,}024, and pruning reduces the width of the MLP from 3{,}584 to 1{,}792 and the vocabulary to 70{,}000 entries.
The model has 368{,}930{,}444 parameters, of which 368{,}570{,}368 are the weights of 187 linear matrices.
Of its 24 layers, 18 are Gated DeltaNet layers (the recurrent layers) and six are gated-attention layers, so the model is a hybrid.
The attention layers are at positions 3, 7, 11, 15, 19 and 23, where the first layer has position 0.
Each recurrent layer has 11 heads of dimension 128 for the query, the key and the value, and a convolution of length~4.
We distilled the model and then adapted it on question--answer examples built on the UT Campus Object Dataset (CODa)~\cite{zhang2024coda}.
The resulting weight file has not been released, and the technical supplement records the training budget, the size of the file and its SHA-256 digest.

\subsection{Number formats}
\label{sec:formats}

\looseness=-1 \emph{Weights.} The 187 linear matrices are stored as 4-bit integer (INT4) codes in groups of 64, and each group has one scale $\sigma$ in the 16-bit floating-point format bfloat16 (bf16).
The quantization is symmetric, with all zero points equal to zero, so the hardware uses codes from $-8$ to $7$.
A codebook stores the dequantized value $T[c]=\mathrm{RNE}_{\mathrm{bf16}}(c\cdot\sigma)$ for each code $c$, where $\mathrm{RNE}_{\mathrm{bf16}}$ rounds to bf16 with ties to even.
The hardware reads the value from the codebook, so it does not multiply by the scale at run time, and this lookup matches the dequantization of the reference for all 368{,}570{,}368 weights.
The parameters of the normalizations, of the convolution and of the decay remain in floating point.
One recurrent layer stores eight matrices, which are the four matrices of the input projections, the matrix of the output projection and the three matrices of the gated MLP, and together they contain 12{,}736{,}512 weights.
The final projection of the model onto the vocabulary, which we refer to as the vocabulary projection, shares its weights with the input embedding.

\looseness=-1 \emph{Activations.} The activations are in bf16.
For the matrix-vector products, they are converted in groups of 64 to a block-floating format, in which the values of a group share one exponent and each value has a mantissa of 19 bits.

\looseness=-1 \emph{State.} Inside the tile, the state of a head is also in a block-floating format, with an 18-bit mantissa $m$ per element and one exponent $e$ per head, so that $S=m\,2^{e-15}$.
In external memory, the mantissas have 16 bits.
When the tile writes a head to memory (page-out), it computes $m_{16}=\mathrm{sat}_{16}(\lfloor(m_{18}+2)/4\rfloor)$ and stores the exponent $e_{16}=e+2$, which compensates for the division by four.
When the tile reads the head back (page-in), it restores $m_{18}=4m_{16}$ and $e=e_{16}-2$.
The conversion loses two bits of the mantissa, and we refer to it as \emph{page rounding}.

\looseness=-1 \emph{Layout in memory.} A \emph{beat} is one 512-bit transfer to or from the memory.
Each head starts with a header beat, whose first byte holds the exponent.
Each data beat stores $2W$ mantissas of 16 bits, where $W$ is the number of lanes of the state engine (Section~\ref{sec:slowrate}-\ref{sec:engines}), and its remaining $512-32W$ bits are zero padding.
A head therefore takes $1+128(128/W)/2$ beats.
The tile transfers the state in \emph{pages} of two heads.

\subsection{Memory traffic and operation count}
\label{sec:traffic}

\looseness=-1 The recurrent state of the 18 recurrent layers has $18\times11=198$ heads of $128\times128$ elements.
At two bytes per element, it occupies 6{,}488{,}064 bytes, which is 6.49~MB or 6.2~MiB.
Table~\ref{tab:traffic} lists the memory traffic that the whole model generates when it decodes one token, by component and for two number formats of the vocabulary projection.
The vocabulary projection is outside the scope of the tile, so the table gives both formats.
All values are calculated from the dimensions of the model, for a context of 744 tokens, and they exclude protocol overhead.
One read and one write of the state without padding would move twice its size, 12.98~MB, as the last row of the table shows.
With the implemented layout ($W=8$), half of every data beat is padding, and one read and one write of all 198 heads move 25{,}977{,}600 bytes, or 25.98~MB.
Even with the padding, the state is 7.7\,\% to 11.2\,\% of the total traffic, and the weights of the layers and of the vocabulary projection account for most of it.
In arithmetic operations, the whole model needs 392{,}015{,}872 MACs per decoded token, of which the recurrent kernel contributes 9{,}732{,}096, or 2.48\,\%.

\begin{table}[t]
\caption{Calculated memory traffic per decoded token for the whole model, in MB, at a context of 744 tokens. The two columns differ in the number format of the vocabulary projection. The last row gives the state traffic without headers and padding, for comparison.}
\label{tab:traffic}
\centering\footnotesize
\begin{tabular}{@{}lrr@{}}
\toprule
 & \multicolumn{2}{c}{Vocabulary projection in} \\
Component & bf16 & INT4 \\
\midrule
INT4 weights of the layers & 148.45 & 148.45 \\
Scales of the weights & 9.28 & 9.28 \\
Vocabulary projection & 143.36 & 38.08 \\
KV-cache reads of the attention layers & 9.14 & 9.14 \\
State read and write, implemented & 25.98 & 25.98 \\
Convolution history & 0.91 & 0.91 \\
Read of one embedding row & 0.002 & 0.002 \\
\midrule
Total & 337.12 & 231.84 \\
Share of the state & 7.7\,\% & 11.2\,\% \\
\midrule
State read and write, without padding & 12.98 & 12.98 \\
\bottomrule
\end{tabular}
\end{table}

\subsection{Engines}
\label{sec:engines}

\looseness=-1 \emph{State engine.} The state engine computes the recurrent kernel.
It has $W$ lanes and processes $W$ state elements per cycle in each pass.
It stores its operands in block memories, not in registers, and the read pass and the write pass share one pair of multipliers per lane.

\looseness=-1 \emph{Matrix-vector engine.} This engine computes all products of a weight matrix and an activation vector, that is, the projections and the MLP.
It has $N_L$ lanes, and each lane multiplies an INT4 weight by a 19-bit mantissa and accumulates the product.
A reader supplies the weights from external memory in bursts of several beats over a 512-bit port, and a floating-point accumulator aligns the sum of each group to a running exponent.

\looseness=-1 \emph{State pager.} The state pager moves the pages between the state engine and external memory over an Avalon memory-mapped (Avalon-MM) interface.
It uses double buffering.
With a simulation model of the memory, the page transfers therefore overlap the computation, and only the first page-in and the last page-out add latency.

\looseness=-1 \emph{Control datapath.} All other operations, namely the normalizations, the convolution with its activation, the scalars $\alpha$ and $\beta$, the gates and the residual additions, run on one shared fp32 datapath.
It has a multiplier, an adder and units for $\exp$, reciprocal, reciprocal square root and $\log(1+x)$, and a short sequence of steps drives it for each operation.
A version with separate arithmetic units for all steps would have needed about 20k adaptive logic modules (ALMs), which is 43\,\% of the device.

\subsection{Bit-exact verification}
\label{sec:verify}

\looseness=-1 Because the state is recurrent, a numerical discrepancy at one token propagates to the later tokens.
We therefore test two properties separately, namely the accuracy of the fixed-point arithmetic against fp32 and the exact agreement between the RTL and the fixed-point reference.
Before the board experiments, the RTL passed five simulation tests against the reference, which the supplement lists.
They cover the state engine on layer~0 (Test~1), the matrix-vector engine on 2{,}992 matrix-vector products with 5{,}304{,}256 output values, the state pager, the mode controller and the whole tile.
The rest of this subsection first measures the approximation error, then describes a defect that Test~1 did not expose, and finally repeats the exact comparison of the state engine on all 18 recurrent layers.
The numerical tests of this subsection use the two prompts (Section~\ref{sec:method}), and Section~\ref{sec:board} reports the tests on the board.

\emph{Approximation error of the state engine.}
The first experiment measures, in software, the error of the fixed-point state engine against fp32.
A fixed-point model of the engine receives the fp32 inputs of the kernel, quantizes them once and then computes 64 tokens.
The engine also limits the number of bits by which the exponent of the state may grow in one token.
We refer to this limit as the \emph{shift limit}, and the parameter \texttt{SH\_MAX} sets it.
The fixed-point reference uses $\texttt{SH\_MAX}=31$, whereas the original RTL used $\texttt{SH\_MAX}=3$, and we evaluate both values.
For each prompt and each recurrent layer, we compute the relative error
\begin{equation}
 E_X=\frac{\max_{t,h,i}|X_{\rm fx}-X_{\rm fp32}|}{\max_{t,h,i}|X_{\rm fp32}|},\quad X\in\{o,S\}.
\label{eq:errglobal}
\end{equation}
Here the subscripts fx and fp32 denote the fixed-point and the fp32 values, $i$ indexes the elements of the output $o$ or of the state $S$, and each maximum spans the 64 tokens $t$ and the 11 heads $h$.
Table~\ref{tab:errorpolicy} reports the largest $E_X$ over the 18 recurrent layers and the two prompts, for both shift limits, with and without page rounding.
With a shift limit of 31, the errors stay below 0.7\,\%.
With a shift limit of 3, they reach 5\,\% for the output and 19\,\% for the state, and page rounding makes no difference to these maxima.
The supplement reports two further measures, namely the error normalized separately for each head and each token, and the error of the whole layer, which is of the order of $10^{-3}$.

\begin{table}[t]
\caption{Relative errors of the fixed-point state engine against fp32, computed with (\ref{eq:errglobal}). Every row covers 18 recurrent layers, two prompts and 64 tokens. Page rounding, where used, is applied at initialization and after each token. These are software results.}
\label{tab:errorpolicy}
\centering\footnotesize
\begin{tabular}{@{}llrr@{}}
\toprule
Shift limit (\texttt{SH\_MAX}) & Page rounding & Max $E_o$ & Max $E_S$ \\
\midrule
31 (reference) & no & $4.69\!\times\!10^{-4}$ & $4.40\!\times\!10^{-4}$ \\
3 (original RTL) & no & $5.15\!\times\!10^{-2}$ & $1.91\!\times\!10^{-1}$ \\
31 (reference) & yes & $6.43\!\times\!10^{-3}$ & $1.10\!\times\!10^{-3}$ \\
3 (original RTL) & yes & $5.15\!\times\!10^{-2}$ & $1.91\!\times\!10^{-1}$ \\
\bottomrule
\end{tabular}
\end{table}

\looseness=-1 \emph{Defect in the shift limit.}
The large errors for a shift limit of 3 in Table~\ref{tab:errorpolicy} come from two events.
We tested both shift limits on 25{,}344 head-tokens each, which are 18 layers, 11 heads and 64 tokens for each of the two prompts.
With a limit of 3, the required growth exceeds the limit twice, at recurrent layer~1, head~1 and token~45, once for each prompt, where all indices start at zero.
In each event, the engine needs to raise the exponent by four bits but may raise it only by three, so one element of the update saturates.
The original RTL counted saturation only in the write pass, whereas this saturation occurs before the write pass, so its counter stayed at zero.
Because the state is recurrent, the error persists into the later tokens and affects many more than two outputs.
For the other 34 combinations of prompt and layer, the limit has no effect on these inputs.

\looseness=-1 \emph{Repair and tests on all layers.}
The repaired RTL uses a shift limit of 31, as the reference does, while the exponent itself remains limited to $-12\leq e\leq8$.
New counters also report saturation before the write pass and each time a limit is reached.
We then compared the repaired state engine with the reference in 72 simulation runs, which cover the 18 recurrent layers, the two prompts and the widths $W=16$ and $W=8$.
All 72 runs pass.
Each run has 64 tokens and compares 90{,}112 output words and 11{,}534{,}336 state mantissas, together with the exponents and the event records.
Test~1 sampled the state at only eight tokens and missed the defect, whereas these runs check the state after every token and therefore include token~45.
For comparison, we also ran the original RTL on recurrent layer~1, and both runs fail, with 43{,}592 and 43{,}382 state mismatches for the two prompts, respectively.

\section{Platform and Methodology}\label{sec:method}

\looseness=-1 All hardware targets one device, an Altera Agilex~5 E-Series SoC-FPGA on the DK-A5E013BM16AEA development kit, which provides two LPDDR4 interfaces and on-board JTAG.
Fig.~\ref{fig:app}(a) shows the kit, Fig.~\ref{fig:app}(b) shows where each part runs, and Fig.~\ref{fig:sys} shows the system as the RTL instantiates it.
All datapaths are in the programmable logic and run without an operating system, so the measurements reflect the architecture alone.
We simulate the tile in Questa Altera Starter FPGA Edition 2025.3 and the engines of the front-end in Vivado 2025.1 XSim, and we build the final bitstreams with Quartus Prime Pro 26.1.1, Build 130.
One compilation of a design into a bitstream is a \emph{build}.

\looseness=-1 The workload of the front-end is the OVAD dataset of occluded vehicles~\cite{schulz2021hearing}, which contains real 56-channel recordings at blind T-junctions.
We quantize the audio and replay it from on-chip memory.

\looseness=-1 The workload of the decode tile is one recurrent layer of the model (Section~\ref{sec:slowrate}).
The numerical tests use two prompts from the CODa question set, namely a counting question and a direction question.
After the scene is serialized into text and the answer options are formatted, they contain 744 and 244 tokens, respectively, and the first prompt sets the context of 744 tokens in Table~\ref{tab:traffic}.
We tokenized the prompts with the public Qwen3.5 tokenizer and mapped the tokens to the 70{,}000-entry vocabulary with a stored table.

\looseness=-1 In simulation, we compare the outputs of the engines with the fixed-point reference.
On the board, test logic compares selected results with stored expected values, and the host reads back counters and signatures through virtual JTAG.
A \emph{signature} is a checksum that the hardware accumulates over a sequence of words.

\section{Front-End Results and Design Study of the Tile}\label{sec:results}

\looseness=-1 This section reports the results of the front-end on the board and the design study of the decode tile, which uses simulation and tile-only builds.
Section~\ref{sec:board} reports the board runs of the tile.

\subsection{Results of the front-end: engines, cache and classifier}

\looseness=-1 We measured the front-end on the board at a clock of 100~MHz and read the results over virtual JTAG.
In six separate test builds, every result word that we read from the device matched the RTL simulation.
The front-end processes one audio frame of the 56 channels in 7.02~ms (702{,}387 cycles), and all 120 outputs on the chip match the reference.

\begin{figure}[t]
\centering
\includegraphics[width=0.60\columnwidth]{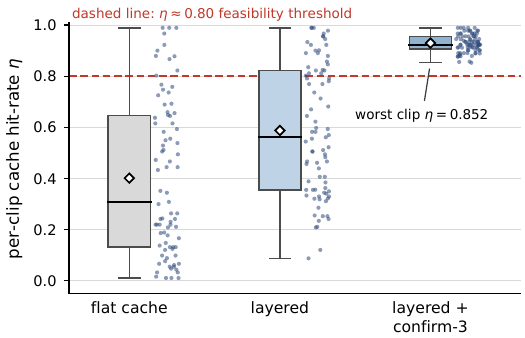}
\caption{Cache hit rate $\eta$ of each of the 83 clips for three versions of the cache policy. Boxes span the quartiles, and whiskers span the full range. Diamonds mark the means, and points mark the individual clips. The dashed line is the feasibility threshold $\eta\approx0.80$ (Section~\ref{sec:invariants}). Only the full policy places every clip above the threshold, and its lowest clip is at $\eta=0.852$.}
\label{fig:eta}
\end{figure}

\looseness=-1 A replay of one recorded clip on the board gives a hit rate of $\eta=0.901$, with 9 misses in 91 hops.
Fig.~\ref{fig:eta} shows the distribution of $\eta$ over the 83 clips of the software replay for three versions of the cache policy.
If the system queries the semantic layer on every change of the scene description, the mean $\eta$ is 0.402 (flat cache in Fig.~\ref{fig:eta}).
If updates of positions are absorbed at the sensor rate and cause no query, the mean $\eta$ is 0.588 (layered cache).
The full policy adds a rule, referred to as \emph{confirm-3}, under which a change of object class or of the acoustic alarm must persist for three consecutive hops before it causes a query.
This rule raises the mean $\eta$ to 0.929 and keeps every clip above the feasibility threshold of 0.80.
The board replay and the software replay implement this rule slightly differently, and on the clip of the board replay they give 9 and 10 misses in 91 hops, respectively, as the supplement explains.

\looseness=-1 The acoustic detection also allows a \emph{prefetch}.
In the clip that we replayed on the board, the front-end detects the occluded vehicle 2.3~s before it becomes visible.
The system can therefore query the semantic layer on the acoustic cue, so that the result is in the cache before the vehicle appears.
In a board test that emulated a latency of the semantic layer of 1{,}500 cycles, the prefetch reduced the blind window from 1{,}500 cycles to zero, and the supplement describes the retained record of this test.
In the software replay of all clips, with emulated latencies of 50, 150 and 500~ms, the result is in the cache when the vehicle becomes visible in 35, 33 and 28 of the 42 events in which a vehicle emerges, and the mean blind window shrinks by 83, 82 and 75\,\%, respectively.

\looseness=-1 The front-end also includes a convolutional neural network (CNN) with INT8 arithmetic, which classifies the direction from which a vehicle approaches from SRP spectrograms of $16\times30$ values.
Its test accuracy on held-out recordings is 80.3\,\% in both floating point and INT8, and on the device its 32 output values for eight held-out segments match the reference exactly.
One inference requires 357{,}760 MACs and takes 0.459~ms.

\subsection{Design study of the tile}
\label{sec:slowres}

\begin{table}[t]
\caption{The two tile configurations that fit the device, from the design study. $N_L$ and $W$ are the numbers of lanes of the matrix-vector engine and of the state engine. Cycle counts come from RTL simulation with the memory model, and times are calculated at 150~MHz. ALM, M20K and DSP are the logic, memory and multiplier blocks of the device, and $F_{\max}$ is the maximum clock frequency. A third configuration, with $N_L=128$ and $W=16$, does not fit the device.}
\label{tab:point}
\centering
\footnotesize
\begin{tabular}{@{}lrr@{}}
\toprule
 & $N_L{=}64$, $W{=}16$ & $N_L{=}128$, $W{=}8$ (chosen) \\
\midrule
Cycles per layer-token & 446{,}100 & 393{,}935 \\
Time at 150\,MHz       & 2.97\,ms  & 2.63\,ms \\
ALM                    & 38{,}888 (83\,\%) & 27{,}105 (58\,\%) \\
M20K                   & 234 (65\,\%) & 250 (70\,\%) \\
DSP                    & 88 (23\,\%)  & 112 (30\,\%) \\
$F_{\max}$             & 155.6\,MHz & 158.8\,MHz \\
\bottomrule
\end{tabular}
\end{table}

\looseness=-1 Before the board experiments, we compared configurations of the tile in RTL simulation with a memory model and in tile-only builds.
Table~\ref{tab:point} lists the two configurations that fit the device.
The chosen configuration has $N_L=128$ lanes in the matrix-vector engine and $W=8$ lanes in the state engine, whereas the alternative has $N_L=64$ and $W=16$.
Halving the state engine costs 29{,}660 cycles per layer-token, but doubling the matrix-vector engine saves 81{,}825, so the chosen configuration is faster.
It also uses fewer ALMs, which are the limiting resource, and leaves 42\,\% of the device for the memory interface and the readback logic.
Of this difference, 2{,}697 ALMs come from the numbers of lanes and the rest from how the RTL describes one memory.

\looseness=-1 The study gave two further results, which the supplement documents.
First, the area depends strongly on how the RTL describes the memory for the converted activations.
Describing this memory as $N_L/8$ independent banks instead of one wide register reduced the tile from 36{,}191 to 27{,}105 ALMs and raised its maximum clock frequency $F_{\max}$ from 154.6 to 158.8\,MHz, without any change of the arithmetic.
Second, sharing the multipliers among the engines does not pay off on this device.
In a test on the 64-lane tile, sharing reduces the DSP blocks from 88 to 72, but it increases the ALMs from 38{,}724 to 39{,}690 and lowers $F_{\max}$ from 161.2 to 159.6~MHz, and logic, not DSP blocks, is the limiting resource.

\subsection{Scheduling of the control datapath and the limit of the memory link}

\looseness=-1 The control datapath processes one entry of a vector at a time, and each of its operations is a chain of dependent steps.
The convolution, for example, needs 50 cycles per entry, and in 43 of them one arithmetic unit waits for another.
Consecutive entries are independent of each other, however, so we overlap them with modulo scheduling, in which a new entry starts after a fixed number of cycles (22 for the convolution).
This change shortens the control datapath from 428{,}306 to 206{,}283 cycles per layer-token, which is $2.08\times$ faster, and it made the layer-token $1.59\times$ faster at the time of this change.
It costs 391 more ALMs and lowers $F_{\max}$ by 5.6\,MHz.
The arithmetic did not change, and the five simulation tests remained exact.
The supplement gives the rules of the schedule and the values for the three largest operations.

\looseness=-1 The matrix-vector phases, in contrast, are limited by the delivery of the weights because the 128-lane engine already uses almost every weight that the memory delivers.
In simulation, the memory model delivers a beat in six of every seven cycles, and the 128-lane engine consumes 99.9\,\% of the delivered beats when it runs alone and 99.8\,\% within the tile.
Each beat carries the weights for 120 MACs out of a possible 128 because the remaining bits hold the scales of the groups.
With 64 lanes, the engine is limited by its lanes and reaches 61.9 of 64 MACs per cycle.
Both configurations therefore operate close to the limit of their matrix-vector engines.
For the chosen configuration, a faster link to the memory or a more compact weight format could shorten these phases.

\section{Board Results of the Decode Tile}
\label{sec:board}

\subsection{Configurations and method}

\looseness=-1 This section reports the board experiments with the decode tile and the LPDDR4 memory of the kit.
We test two system configurations, P and M.
Configuration P contains the tile alone, and configuration M is the complete system (Fig.~\ref{fig:sys}), in which the tile is one node of the $2\times3$ mesh, together with five processing elements and a scheduler in the programmable logic.
Both configurations use the eight-lane state engine.
Memory bridges connect the tile to the LPDDR4 controller (Fig.~\ref{fig:sys}) and transfer bursts of eight beats.
For each configuration, we compiled three builds with different placement seeds, where a \emph{seed} initializes the placement algorithm of the tool, so that different seeds give different implementations.
Table~\ref{tab:silicon} lists the resources and the timing of these six builds.
The resources include the memory interface, the readback logic and the logs, so they are not directly comparable with the tile-only values in Table~\ref{tab:point}.
All builds run at 75~MHz, which is below their $F_{\max}$ of 98.56 to 110.66~MHz.

\begin{table}[t]
\caption{The six final board builds. P is the tile alone, and M is the complete system (Fig.~\ref{fig:sys}). All builds run at 75~MHz. The ranges span the three placement seeds. $F_{\max}$ is the reported value for the clock of the tile.}
\label{tab:silicon}
\centering\footnotesize
\begin{tabular}{@{}lrrrr@{}}
\toprule
Config. & ALMs & M20Ks & DSPs & $F_{\max}$ (MHz) \\
\midrule
P & 29{,}958--29{,}979 & 298 & 108 & 102.64--110.66 \\
M & 40{,}025--40{,}077 & 305 & 128 & 98.56--108.77 \\
\bottomrule
\end{tabular}
\end{table}

\looseness=-1 We programmed each build five times, that is, we loaded its bitstream into the device five times, and we started one run of 64 layer-tokens after each programming, which gives 30 runs.
In these runs, the complete layer uses a synthetic weight image of 6.8~MB in place of the trained weights.
The image has the addresses and the sizes of the real weights, and its content is a fixed hash of the address.
The simulation generates the same image, so every data value that the board produces has an exact expected value.
Latency counters depend on the memory and are reported as measurements.
The recurrent state is kept from token to token.
The hardware logs one \emph{record} for each layer-token in P and for each pair of layer-tokens in M, where the scheduler runs the layer-tokens in pairs.
A record contains signatures, counters and status fields.
The host reads the records through virtual JTAG, and at selected tokens it also reads the complete output and the complete state pages.

\subsection{Exact comparisons}

\looseness=-1 \emph{Kernel test with real inputs.}
A separate board design tests the recurrent kernel with real input vectors from recurrent layers 0, 1 and 10 and from both prompts.
Each of these six cases runs twice over 64 tokens, and all 12 runs pass.
After each token, a comparator on the device checks the complete output, the packed state pages, the exponents in the page headers and the event counters.
To confirm that the comparator detects errors, we flipped one bit in the expected state, and the comparator reported it.
This test checks the 16-bit pages as they are stored in memory, whereas the simulations (Section~\ref{sec:slowrate}-\ref{sec:verify}) check the internal 18-bit mantissas.

\begin{table}[t]
\caption{Results of the three board tests of the complete layer, for the configurations P and M.}
\label{tab:meshchecks}
\centering\footnotesize
\begin{tabular}{@{}lrrr@{}}
\toprule
Test & P passes & M passes & Layer-tokens \\
\midrule
Runs after programming & 15/15 & 15/15 & 1{,}920 \\
Runs of 1{,}024 layer-tokens & 1/1 & 1/1 & 2{,}048 \\
Initial run and ten soft resets & 11/11 & 11/11 & 352 \\
\bottomrule
\end{tabular}
\end{table}

\looseness=-1 \emph{Tests of the complete layer.}
Table~\ref{tab:meshchecks} summarizes three board tests of the complete layer with the synthetic weights, and every run passes.
The first board test consists of the 30 runs (Section~\ref{sec:board}-A), and all of them pass.
With 64 layer-tokens per run, they contain 1{,}920 layer-tokens, and 47{,}520 record fields agree exactly with the simulation.
A record field is a signature, a counter or a status value, not an individual word of a tensor.
In addition, the complete output and the complete state pages are read back and compared word by word at selected tokens, which the supplement lists, and at the last token of every run.
Four of the five runs of each build execute without interruption, and the fifth pauses for detailed readback, so we exclude it from the latency statistics.
In the second board test, P and M each pass one uninterrupted run of 1{,}024 layer-tokens, during which the recurrent state is never reset or reloaded.
In the third board test, each configuration passes eleven runs of 16 layer-tokens, namely an initial run and ten runs that each follow a soft reset from the idle state, that is, a reset through a control register without reprogramming.
In total, the three tests comprise 54 runs, and all 106{,}920 comparisons of record fields are exact.

\subsection{Latency and the cost of paging}

\looseness=-1 This subsection answers the first research question of the paper.
In P, the 12 uninterrupted runs give 768 records, one per layer-token.
The median latency of a layer-token is 836{,}555 cycles, or 11.154~ms at 75~MHz, with a range of 836{,}249 to 837{,}124 cycles.
Of this latency, the matrix-vector phases take a median of 516{,}318 cycles (62\,\%), the control phases take 206{,}283 cycles (25\,\%), and the state phase takes a median of 111{,}817 cycles (13\,\%).
The state phase includes the update of the state, the paging and a wait of 187 cycles for the write responses of the memory (Section~\ref{sec:board}-D).
The state phase is thus the shortest of the three, and the matrix-vector phases, which depend on the delivery of the weights, are the longest.
Paging the state is therefore not the bottleneck of this layer.

\begin{figure}[t]
\centering
\begin{tikzpicture}
\begin{loglogaxis}[width=0.98\columnwidth, height=3.5cm, font=\scriptsize,
  xlabel={weight bytes delivered per cycle}, ylabel={MAC per cycle},
  xmin=8, xmax=160, ymin=12, ymax=200,
  xtick={10,20,50,100}, xticklabels={10,20,50,100}, ytick={16,32,64,128}, yticklabels={16,32,64,128},
  grid=major, grid style={black!12}, tick style={black!40}]
\addplot[black, thick, domain=8:68.3, samples=2] {1.8741*x};
\addplot[black, thick, domain=68.3:160, samples=2] {128};
\addplot[only marks, mark=*, mark size=2.2pt, black] coordinates {(13.16,24.67) (29.16,54.66)};
\addplot[only marks, mark=o, mark size=2.2pt, black, thick] coordinates {(54.81,102.71)};
\addplot[black!60, dashed, thick] coordinates {(56.87,12) (56.87,128)};
\addplot[black!60, dashed, thick, forget plot] coordinates {(113.73,12) (113.73,128)};
\node[font=\scriptsize, anchor=south west, rotate=25, inner sep=1pt] at (axis cs:9.6,18.5) {link roof: 1.874 MAC/byte};
\node[font=\scriptsize, anchor=south east, inner sep=1pt] at (axis cs:152,131) {compute roof: 128 lanes};
\node[font=\scriptsize, anchor=north west] at (axis cs:13.16,24.67) {\ 8 beats};
\node[font=\scriptsize, anchor=north west] at (axis cs:29.16,54.66) {\ 128 beats};
\node[font=\scriptsize, anchor=north west, xshift=3pt, yshift=-1pt] at (axis cs:54.81,102.71) {model};
\node[font=\scriptsize, anchor=south east, inner sep=1pt] at (axis cs:55.3,12.6) {150\,MHz};
\node[font=\scriptsize, anchor=south east, inner sep=1pt] at (axis cs:110.5,12.6) {75\,MHz};
\end{loglogaxis}
\end{tikzpicture}
\caption{Roofline plot of the memory link for the weights, from the experiment with the state on chip. It shows the rate of MACs as a function of the rate at which weight bytes arrive. The layer performs 12{,}736{,}512 MACs over a weight image of 6{,}796{,}032 bytes, which gives 1.874 MACs per byte (sloped line). The 128 lanes limit the rate to 128 MACs per cycle (horizontal line), and the two lines meet at 68.3 bytes per cycle. Filled markers are measurements on the device with bursts of 8 and 128 beats, and the open marker is the memory model. Dashed lines show the peak rate of the LPDDR4 memory, 8.53~GB/s, at 75 and 150~MHz.}
\label{fig:roof}
\end{figure}
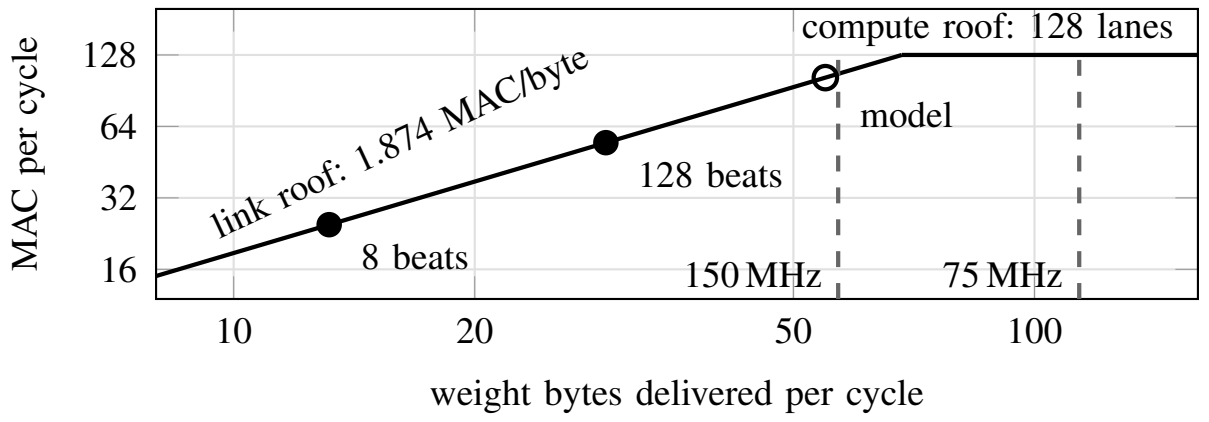

\looseness=-1 This latency is larger than 2.63~ms (Table~\ref{tab:point}), mainly because the board runs at 75~MHz instead of 150~MHz and because the memory bridge of the board delivers the weights more slowly than the memory model of the simulation.
An earlier build, which kept the state of its one layer on chip and read the same synthetic weights from the LPDDR4 memory, shows the size of the second effect.
With bursts of eight beats, a layer-token took 786{,}220 cycles in this build, whereas it takes 393{,}935 cycles in simulation with the memory model, and all of the increase is in the matrix-vector phases.
The bridge delivered 0.21 beats per cycle, which is about 1.0~GB/s at 75~MHz.
With bursts of 128 beats, the latency fell to 502{,}897 cycles, or 6.7~ms.
Fig.~\ref{fig:roof} is a roofline plot of these results, that is, a plot of the rate of MACs against the rate at which the weight bytes arrive over the link from the memory, with the limits that this memory link and the engine impose.
Both measurements lie on the sloped line, where the memory link limits the rate, and not on the horizontal line, where the 128 lanes would limit it.
The final tile also uses bursts of eight beats, and the 516{,}318 cycles of its matrix-vector phases correspond to the lower of the two measured points.
The build with 128-beat bursts predates the repairs (Section~\ref{sec:board}-D), so the 6.7~ms belong to that earlier build and not to the final tile.

\looseness=-1 In M, the 12 uninterrupted runs give 384 records, one per pair of layer-tokens.
The median duration of a pair is 2{,}234{,}305 cycles, or 29.791~ms.
This duration includes the loading of the tile through the mesh, the routing and the work of the scheduler, so it is not the latency of two layer-tokens.

\subsection{Implementation repairs}

\looseness=-1 The six final builds (Table~\ref{tab:silicon}) include four repairs because the original builds failed some board runs although they had met their timing constraints.
In the original build of the tile alone, for example, the first token was correct in all five runs, but the second token was correct in only three.
The repairs are as follows.
First, we set the shift limit of the RTL to that of the reference (Section~\ref{sec:slowrate}-\ref{sec:verify}).
Second, we changed the ready signal of the port of the processing element, which removed five combinational loops.
Third, the pager now waits until the memory has acknowledged all pending writes before it reads a page that depends on them.
Fourth, we added registers that hold the operands of the control datapath stable while its arithmetic units evaluate them over several cycles, and the timing exceptions that allow these units more than one clock cycle now start at these registers.
The fourth repair was necessary because builds with only the first three repairs failed all 15 runs of P and the five runs of one seed of M, although their timing reports showed positive slack.
With all four repairs, the six final builds pass every test in Table~\ref{tab:meshchecks}, and they meet all timing requirements that the tool analyzes at 75~MHz, with a setup slack of 3.590 to 4.296~ns and a hold slack of 0.000 to 0.005~ns.
This evidence supports the four repairs as a set, and the supplement documents the original failures and the checks of each repair.

\section{Discussion, Limitations, and Conclusion}
\label{sec:discussion}

\subsection{Interpretation and limitations}

\looseness=-1 The evidence supports paging as a practical way to store the state of this model when it decodes one sequence at a time.
This conclusion depends on the dimensions of the model, the page layout, the burst behavior of the memory and the batch size, so it should not be extended to models with a larger state or to batched workloads without new measurements.

\looseness=-1 The evidence has the following limits.
Exact agreement holds for the tested interfaces and inputs, that is, for two prompts and for one device under nominal conditions, and it does not establish the accuracy of the model on a task.
For the complete layer, which runs with synthetic weights on the board, the readback of the complete state covers selected tokens, and signatures cover the other tokens.
We did not evaluate longer contexts, and we did not implement the six attention layers, the vocabulary projection or the processing of the prompt.
The front-end and the tile have not yet run at the same time, and the forwarding of flits between two systems is shown only within one FPGA.
The supplement records three remaining implementation limits.
We did not formally verify the signals that cross between clock domains, and a processing element drops a result when its output is blocked.
We also did not measure power, and the supplement explains why the estimate of the vendor tool, about 0.35~W of dynamic power for the tile, hardly depends on the workload.

\subsection{Conclusion}

\looseness=-1 We built and measured a decode tile that pages the recurrent state, for one complete Gated DeltaNet layer of a custom 369M model on an edge FPGA.
For the whole model, paging moves 26.0~MB per decoded token, which is 7.7--11.2\,\% of the calculated memory traffic.
Of the 11.15~ms that the tile needs per layer-token at 75~MHz, the matrix-vector phases take about 62\,\% and the state phase about 13\,\%.
The state is thus too large for block memory, but the dominant measured cost is the delivery of the weights and not the paging.
The tile operates as one node of a multicast mesh of tri-mode processing elements, whose mode switch costs 3 cycles from idle in simulation and at most 34 cycles on the device.

\looseness=-1 The verification flow separates the numerical approximation from the exact agreement of the implementation with its reference.
Comparisons of the state on all 18 recurrent layers exposed a wrong shift limit, and the repaired RTL passes all 72 simulation runs of the state engine.
On the board, the six final builds pass all 30 runs after programming, as well as the tests with resets and with long sequences.
These results establish exact agreement at the points of comparison.

\subsection*{Supplementary material}
A technical supplement specifies the numerical formats, the test coverage, the build records, the design study and the original failures.
An evidence archive contains the RTL, the scripts, the reference records and the board logs, but not the trained weights or some of the generated test vectors.

\section*{Acknowledgment}
OpenAI Codex and Anthropic Claude assisted with language editing and organization of the abstract, Sections I--IX, and the technical supplement.
Under the authors' direction, they also assisted with the test scripts, the repairs of Section~\ref{sec:board}-D, and the log cross-checks.

\makeatletter\let\v@origneedspace\@IEEEtranneedspace
\def\@IEEEtranneedspace#1#2{\global\let\@IEEEtranneedspace\v@origneedspace}\makeatother
\begin{IEEEbiography}[{\includegraphics[width=1in,height=1.25in,clip,keepaspectratio]{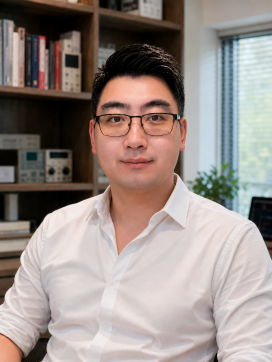}}]{CONG XU}
\looseness-1\ (Member, IEEE) received the B.S. degree in electrical engineering and automation from Zhengzhou University, Zhengzhou, China, and the M.S. degree in electrical engineering from the University of South Florida, Tampa, FL, USA, in 2020, where he is currently pursuing the Ph.D. degree in electrical engineering with the Department of Electrical and Computer Engineering, as a member of the Interdisciplinary Communications, Networking and Signal Processing (iCONS) Lab.

\looseness-1 His doctoral research asks how language-model reasoning can be brought onto resource-constrained embodied platforms, spanning perception algorithms, language-model interfaces and hardware implementations. His research interests include language--vision collaborative SLAM, structured world models for embodied agents, cross-modal risk perception under sensor degradation, semantic serialization as a perception interface for small language models, and efficient on-device AI on FPGA and SoC platforms. He is currently a Technology Specialist with Global Electronics Testing Services, Tampa, FL, USA.
\end{IEEEbiography}

\begin{IEEEbiography}[{\includegraphics[width=1in,height=1.25in,clip,keepaspectratio]{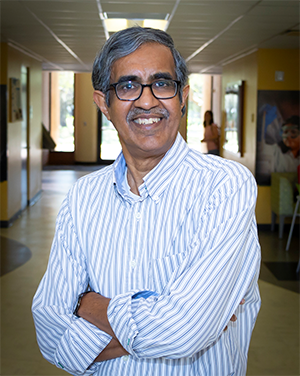}}]{RAVI SANKAR}
\looseness-1\ (Life Senior Member, IEEE) received the B.E. (Honors) degree in electronics and communication engineering from the University of Madras, India, the M.Eng. degree in electrical engineering from Concordia University, Canada, and the Ph.D. degree in electrical engineering from The Pennsylvania State University, USA. He has been with the Department of Electrical and Computer Engineering, University of South Florida, Tampa, FL, since 1985, where he is currently a USF Theodore and Venette Askounes-Ashford Distinguished Scholar award-winning Professor and the Director of the interdisciplinary Communications, Networking and Signal Processing (iCONS) research lab. He was a 2015--16 Fulbright Fellow to Brazil to conduct collaborative research focusing particularly on the use of multiple sensors and signal processing to improve healthcare, and was a visiting research fellow of the Japanese Society for the Promotion of Science (JSPS), nominated by the NSF in spring 2000 to conduct collaborative research in Japan. He also held visiting positions at the University of Melbourne, Australia, in summer 2000, the U.S. Air Force Research Lab (Rome Lab), Rome, NY, in summer 1997, and Motorola, Boynton Beach, FL, in summer 1991.

\looseness-1 Prof.\ Sankar's main research interests are in the areas of wireless communications, networking, signal processing, and their applications. His current focus is on advancing healthcare and intelligent systems through multimodal sensing, signal processing, machine learning and AI, wearable technologies, and edge intelligence. He has published extensively in those areas with over 250 papers in journals and premier international conferences and several book chapters. His research has been cited widely, as measured by an h-index of 33 and an i10-index of 93. Through the years, he has supervised 8 post-doctoral researchers, over 72 Ph.D. and M.S. students, and numerous B.S. senior capstone design projects. Further, he has mentored many more students, including another 15 Ph.D. students, and is currently directing 3 Ph.D. students. The iCONS research lab under his leadership has successfully conducted numerous funded research projects over the years with support from various federal and state agencies and industries. Research contributions of the lab have been widely recognized for their quality and reputation and further advanced by fostering major international collaborations in South Korea and Brazil.

\looseness-1 Prof.\ Sankar was a Distinguished Lecturer for the IEEE Engineering in Medicine and Biology Society (EMBS) in 2014--16. He has delivered numerous (more than 50) plenary, keynote, or invited lectures at international conferences and institutions over the years all over the world, including Korea, Japan, Mexico, Brazil, and India. He has received numerous awards, including the IEEE Florida Council Outstanding Engineering Educator award and the Outstanding Contributions in Research award from the ASEE. He has served on the editorial board of several journals, including as an Associate Editor of \emph{IEEE Communications Surveys and Tutorials}, on organizing committees and technical program committees, and as a session organizer and chair for many flagship IEEE conferences. He was the organizer and co-chair of the US--Korea Joint International Workshop on Global Wireless Sensor Networks (GWSN), sponsored by the NSF and KOSEF (Korea Science and Engineering Foundation), held in Korea in 2009 and 2011. He has served the IEEE in various capacities, such as the Founding Chair of the Engineering in Medicine and Biology Society (EMBS) chapter of the IEEE Florida West Coast Section and the Vice-Chair of the IEEE Signal Processing Society chapter.
\end{IEEEbiography}


\begin{thebibliography}{99}
\bibitem{katharopoulos2020linear} A. Katharopoulos \emph{et al.}, ``Transformers are
RNNs: Fast autoregressive transformers with linear attention,'' in \emph{Proc.
ICML}, 2020, pp. 5156--5165.
\bibitem{yang2025gdn} S. Yang, J. Kautz, and A. Hatamizadeh, ``Gated delta networks: Improving
Mamba2 with delta rule,'' in \emph{Proc. ICLR}, 2025.
\bibitem{schulz2021hearing} Y. Schulz, A. K. Mattar, T. M. Hehn, and J. F. P. Kooij, ``Hearing what you cannot see: Acoustic vehicle
detection around corners,'' \emph{IEEE Robot. Autom. Lett.}, vol.~6,
no.~2, pp.~2587--2594, 2021.
\bibitem{xu2026lvcworld} C. Xu and R. Sankar, ``LVC-World: Structured state
abstraction for prediction under perception degradation,'' preprint,
Zenodo, doi:10.5281/zenodo.21317636, 2026.
\bibitem{gupta2026persistent} N. Gupta \emph{et al.}, ``A persistent-state
dataflow accelerator for memory-bound linear attention decode on FPGA,''
arXiv:2603.05931, 2026.
\bibitem{schlag2021fwp} I. Schlag, K. Irie, and J. Schmidhuber, ``Linear transformers are secretly
fast weight programmers,'' in \emph{Proc. ICML}, 2021, pp. 9355--9366.
\bibitem{yang2024gla} S. Yang \emph{et al.}, ``Gated linear attention
transformers with hardware-efficient training,'' in \emph{Proc. ICML}, 2024,
pp. 56501--56523.
\bibitem{dao2024ssd} T. Dao and A. Gu, ``Transformers are SSMs: Generalized models and efficient
algorithms through structured state space duality,'' in \emph{Proc. ICML}, 2024, pp. 10041--10071.
\bibitem{hong2022dfx} S. Hong \emph{et al.}, ``DFX: A
low-latency multi-FPGA appliance for accelerating transformer-based text
generation,'' in \emph{Proc. MICRO}, 2022, pp. 616--630.
\bibitem{chen2024spatial} H. Chen \emph{et al.},
``Understanding the potential of FPGA-based spatial acceleration for large
language model inference,'' \emph{ACM Trans. Reconfigurable Technol. Syst.},
vol. 18, no. 1, pp. 1--29, Mar. 2025, doi:10.1145/3656177.
\bibitem{zeng2024flightllm} S. Zeng \emph{et al.}, ``FlightLLM: Efficient
large language model inference with a complete mapping flow on FPGAs,'' in
\emph{Proc. ACM/SIGDA FPGA}, 2024, pp. 223--234.
\bibitem{tellme2025} Y. Qiao \emph{et al.}, ``TeLLMe v2: An efficient
end-to-end ternary LLM prefill and decode accelerator with table-lookup matmul
on edge FPGAs,'' arXiv:2510.15926, 2025.
\bibitem{edgellm2024} M. Huang \emph{et al.}, ``EdgeLLM: A
highly efficient CPU-FPGA heterogeneous edge accelerator for large language
models,'' \emph{IEEE Trans. Circuits Syst. I, Reg. Papers}, vol. 72, no. 7, pp. 3352--3365, 2025.
\bibitem{emamba2025} J. Kim \emph{et al.},
``eMamba: Efficient acceleration framework for Mamba models in edge
computing,'' in \emph{Proc. CODES+ISSS}, 2025.
\bibitem{mambax2025} D. Yoon \emph{et al.}, ``Mamba-X: An
end-to-end vision Mamba accelerator for edge computing devices,'' in
\emph{Proc. ICCAD}, 2025.
\bibitem{wei2025lightmamba} R. Wei \emph{et al.},
``LightMamba: Efficient Mamba acceleration on FPGA with quantization and
hardware co-design,'' in \emph{Proc. DATE}, 2025.
\bibitem{marca2024} J. Li \emph{et al.}, ``MARCA: Mamba
accelerator with reconfigurable architecture,'' in \emph{Proc. ICCAD},
2024.
\bibitem{zhong2025specmamba} L. Zhong \emph{et al.}, ``SpecMamba: Accelerating Mamba inference on
FPGA with speculative decoding,'' in \emph{Proc. ICCAD}, 2025.
\bibitem{xu2026lowrankssm} H. Xu \emph{et al.}, ``LowRank-SSM: Hardware-software co-design for
rank-reduced Mamba acceleration on FPGA,'' arXiv:2608.02954, 2026.
\bibitem{frantar2023gptq} E. Frantar \emph{et al.}, ``GPTQ: Accurate
post-training quantization for generative pre-trained transformers,'' in
\emph{Proc. ICLR}, 2023.
\bibitem{lin2024awq} J. Lin \emph{et al.}, ``AWQ: Activation-aware weight quantization for on-device
LLM compression and acceleration,'' in \emph{Proc. MLSys}, 2024.
\bibitem{rouhani2020msfp} B. D. Rouhani \emph{et al.}, ``Pushing the limits of narrow precision
inferencing at cloud scale with Microsoft Floating Point,'' in \emph{Proc. NeurIPS},
2020.
\bibitem{madangopal2026precog} A. Madan Gopal \emph{et al.}, ``Structured memory for edge language
models: Persistent context and corpus retrieval via O(1) SSM state injection,''
arXiv:2608.02560, 2026.
\bibitem{lin2026deltalog} J. Lin, J. Sun, and G. Sun, ``DeltaLog: Deferred materialization of
recurrent states for linear attention decoding,'' arXiv:2608.15533, 2026.
\bibitem{zhang2025pdswap} Y. Zhang \emph{et al.}, ``PD-Swap: Prefill-decode logic
swapping for end-to-end LLM inference on edge FPGAs via dynamic partial
reconfiguration,'' arXiv:2512.11550, 2025.
\bibitem{oh2026trine} H. Oh \emph{et al.}, ``TRINE: A token-aware, runtime-adaptive FPGA
inference engine for multimodal AI,'' in \emph{Proc. DAC}, 2026.
\bibitem{colagrande2025mcast} L. Colagrande and L. Benini, ``A multicast-capable AXI crossbar for
many-core machine learning accelerators,'' in \emph{Proc. AICAS}, 2025, pp. 1--5.
\bibitem{ieee754} \emph{IEEE Standard for Floating-Point Arithmetic}, IEEE Std 754-2019, 2019.
\bibitem{hauser_softfloat} J. R. Hauser, ``The HardFloat, SoftFloat, and TestFloat packages,'' accessed Nov. 6, 2026. \url{http://www.jhauser.us/arithmetic/}
\bibitem{buhrer2026bitlogic} S. B\"{u}hrer, A. Plesner, A. Till, and R. Wattenhofer, ``BitLogic:
Training framework for gradient-based FPGA-native neural networks,'' \emph{Trans. Mach. Learn.
Res.}, 2026.
\bibitem{kanani2026duet} A. Kanani, S. Lee, H. Lyu, J. Lin, J. Park, and U. Y. Ogras, ``DUET: Disaggregated hybrid
Mamba-Transformer LLMs with prefill and decode-specific packages,'' in \emph{Proc. DAC}, 2026.
\bibitem{hwang2024emma} J.-J. Hwang \emph{et al.}, ``EMMA: End-to-end
multimodal model for autonomous driving,'' arXiv:2410.23262, 2024.
\bibitem{tian2024drivevlm} X. Tian \emph{et al.}, ``DriveVLM: The
convergence of autonomous driving and large vision-language models,''
arXiv:2402.12289, 2024.
\bibitem{kim2024openvla} M. J. Kim \emph{et al.}, ``OpenVLA: An open-source
vision-language-action model,'' in \emph{Proc. CoRL}, 2024, pp. 2679--2713.
\bibitem{taherin2025} A. Taherin \emph{et al.}, ``Cross-platform scaling
of vision-language-action models from edge to cloud GPUs,''
arXiv:2509.11480, 2025.
\bibitem{wen2024tinyvla} J. Wen \emph{et al.}, ``TinyVLA: Towards fast,
data-efficient vision-language-action models for robotic
manipulation,'' arXiv:2409.12514, 2024.
\bibitem{xu2025vlacache} S. Xu \emph{et al.}, ``VLA-Cache: Efficient vision-language-action manipulation via
adaptive token caching,'' arXiv:2502.02175, 2025.
\bibitem{sun2026unifs} L. Sun \emph{et al.}, ``UniFS: Unified
fast-to-slow hierarchical architecture for vision-language-action
models,'' arXiv:2606.22794, 2026.
\bibitem{williams2026litevla} J. Williams \emph{et al.}, ``LiteVLA-H: Dual-rate vision-language-action inference for
onboard aerial guidance and semantic perception,'' arXiv:2605.00884,
2026.
\bibitem{knapp1976gcc} C. H. Knapp and G. C. Carter, ``The generalized
correlation method for estimation of time delay,'' \emph{IEEE Trans.
Acoust., Speech, Signal Process.}, vol.~24, no.~4, pp.~320--327, 1976.
\bibitem{dibiase2001srp} J. H. DiBiase, H. F. Silverman, and M. S.
Brandstein, ``Robust localization in reverberant rooms,'' in
\emph{Microphone Arrays: Signal Processing Techniques and
Applications}. Springer, 2001, pp.~157--180.
\bibitem{grondin2019odas} F. Grondin and F. Michaud, ``Lightweight and
optimized sound source localization and tracking methods for open and
closed microphone array configurations,'' \emph{Robot. Auton. Syst.}, vol.~113,
pp. 63--80, 2019.
\bibitem{dasilva2017fpga} B. da Silva, A. Braeken, K. Steenhaut, and A. Touhafi, ``Design
considerations when accelerating an FPGA-based digital microphone array
for sound-source localization,'' \emph{J. Sensors}, vol.~2017,
art.~6782176, 2017.
\bibitem{yin2023srpedge} J. Yin and M. Verhelst, ``CNN-based robust
sound source localization with SRP-PHAT for the extreme edge,''
\emph{ACM Trans. Embedded Comput. Syst.}, vol.~22, no.~3, pp.~1--27, 2023, doi:10.1145/3586996.
\bibitem{qwen2026base} Qwen Team, ``Qwen3.5-0.8B-Base,'' Hugging Face model card, 2026. \url{https://huggingface.co/Qwen/Qwen3.5-0.8B-Base}
\bibitem{zhang2024coda} A. Zhang \emph{et al.}, ``Toward robust robot 3-D perception in urban environments: The UT campus object dataset,'' \emph{IEEE Trans. Robot.}, vol. 40, pp. 3322--3340, 2024.
\end{thebibliography}
\end{document}